# On "A Kalman Filter-Based Algorithm for IMU-Camera Calibration: Observability Analysis and Performance Evaluation"

Yuanxin Wu

The above-mentioned work [1] presented an extended Kalman filter for calibrating the misalignment between a camera and an IMU. As one of the main contributions, the locally weakly observable analysis was carried out using Lie derivatives. The seminal paper [1] is undoubtedly the cornerstone of current observability work in SLAM and a number of real SLAM systems have been developed on the observability result of this paper, such as [2, 3]. However, the main observability result of this paper [1] is founded on an incorrect proof and actually cannot be acquired using the local observability technique therein, a fact that is apparently not noticed by the SLAM community over a number of years. In specific, this note points out that the main observability conclusion cannot be drawn from its proof, while the actual conclusion is also incorrect.

For the sake of clarity, the main conclusion in the original paper (Lemma 1, Corollary 1 and Lemma 3, page 1150) is repeated blow.

*Lemma 1: The matrix*

$$A_{ij} = \begin{bmatrix} \Gamma_{ij} & \Upsilon_{ij} \\ 2\bar{q}_I^T & 0_{1\times 3} \end{bmatrix}$$

*formed by the fifth and sixth block-row elements of the first and last block-columns of the observability matrix O [cf. (50)], with $\Gamma_{ij}$ and $\Upsilon_{ij}$ defined in (57), is full rank.*

*Colollary 1: The matrix described by (60) is full rank if the IMU-camera rig is rotated about at least two different axes.*

*Lemma 3: The observability matrix O [cf. (50)] is full rank when the IMU-camera rig is rotated about at least two different axes.*

---

This work was supported in partin part by the Fok Ying Tung Foundation (131061) and the National Natural Science Foundation of China (61174002).
Yuanxin Wu is with School of Aeronautics and Astronautics, Central South University, Changsha, Hunan 410083, P.R. China, E-mail: (yuanx_wu@hotmail.com).

As indicated in the proof (the 2$^{nd}$ sentence of the paragraph above Eq. (61)), the validity of Lemma 3 is founded on Lemma 1 and Corollary 1. Discussions are focused on Lemma 1 and Corollary 1 in the sequel.

Lemma 1 was proved in Appendix I. As defined in (57), $\Gamma_{ij}$ and $\Upsilon_{ij}$ contain only the matrix blocks corresponding to the two nonzero elements of $\omega_m$. So Lemma 1 is valid if at least two elements of $\omega_m$ are nonzero. The statement above (57) "*…in order to prove observability, only two of the elements of $\omega_m$ need to be nonzero…*" conveyed the same meaning. The above sufficient condition of Lemma 1 was improperly omitted at page 1150.

Note that the authors switched to an inconsistent statement below (62) "*…that are excited in order for these rows to be included in the observability matrix…*". From here on, the number of nonzero components was confused with the degrees of freedom. For example, they thought that there are two rotational degrees of freedom in one rotation with $\omega_m = \begin{bmatrix} 1 & 2 & 0 \end{bmatrix}^T$. It is the inconsistent statement that unfortunately leads to the current Corollary 1 and Lemma 3.

So the proven Corollary 1 and Lemma 3 should be putted as:

Corollary 1c: The matrix (60) is full rank if the IMU-camera rig is rotated at an angular velocity with at least two nonzero components.

Lemma 3c: The observability matrix O [cf. (50)] is full rank when the IMU-camera rig is rotated at an angular velocity with at least two nonzero components.

For comparison, the two different sufficient conditions are listed in Table I.

The sufficient condition of Corollary 1 is tighter than that of Corollary 1c. The basic difference is the number of rotations required. In Corollary 1c, one rotation is generally needed unless the rotation is rarely along one of the coordinate axes, but two or more independent and successive rotations are required for Corollary 1.

TABLE I. TWO DIFFERENT SUFFICIENT CONDITIONS

|  | **Corollary 1 and Lemma 3 (non-proven)** | **Corollary 1c and Lemma 3c (proven)** |
|---|---|---|
| **Sufficient condition** | rotated about at least two different axes | rotated at an angular velocity with at least two nonzero components |

It should be highlighted that authors were investigating by Lie derivatives the local observability [4, 5] that is an instantaneous property of the system of interest. It can be applied to the case of instantaneous rotation and but not to the cases involving two or more (independent and successive) rotations because rotations "*about at least two different axes*" cannot be instantaneously performed. The observability matrix in (50) should only consist of Lie derivatives of the system state at some instantaneous time.

Now let us check the correctness of the actual observability conclusion (Corollary 1c and Lemma 3c) that the authors have proven. Consider a rotation about the axis that is parallel to the position vector of camera with respect to IMU, $^I p_C$ (see Fig. 1). Here we assume the common case that $^I p_C$ has no zero elements. According to the proven conclusion (Corollary 1c and Lemma 3c), the system is observable since the angular velocity now has three nonzero components. But the obvious truth is that $^I p_C$ is unobservable because there is a scale ambiguity in $^I p_C$ along the rotating axis. This is a contradiction.

We tried but did not manage to figure out what leads to the contradiction. Hope that interested readers can clarify it.

## REFERENCES

[1] F. M. Mirzaei and S. I. Roumeliotis, "A Kalman Filter-Based Algorithm for IMU-Camera Calibration: Observability Analysis and Performance Evaluation," *IEEE Trans. on Robotics,* vol. 24, pp. 1143-1156, 2008.

[2] S. Weiss, M. Achtelik, S. Lynen, M. Chli, and R. Siegwart, "Real-time Onboard Visual-Inertial State Estimation and Self-Calibration of MAVs in Unknown Environments," in *IEEE ICRA*, 2012.

[3] J. Kelly and G. S. Sukhatme, "Visual–Inertial Sensor Fusion: Localization, Mapping and Sensor-to-Sensor Self-calibration," *The International Journal of Robotics Research,* vol. 30, pp. 56-79, 2011.

[4] R. Hermann and A. J. Krener, "Nonlinear Controllability and Observability," *IEEE Transactions on Automatic Control,* vol. 22, pp. 728-740, 1977.

[5] W. J. Terrell, "Local Observability of Nonlinear Differential-Algebraic Equations (DAEs) From the Linearization Along a Trajectory," *IEEE Transactions on Automatic Control,* vol. 46, pp. 1947-1950, 2001.